\RequirePackage{fix-cm}
\documentclass[twocolumn]{svjour3}          
\smartqed  
\usepackage{graphicx}

\usepackage{amssymb}
\usepackage{amsmath}
\usepackage{latexsym}
\usepackage[]{algorithm2e}
\usepackage{subfig}
\SetKwRepeat{Do}{do}{while}

\usepackage[colorlinks, linkcolor=red, citecolor=blue]{hyperref}
\begin{document}
\RestyleAlgo{boxruled}
\LinesNumbered

\title{Unsupervised Image Segmentation using the Deffuant-Weisbuch Model from Social Dynamics
}


\author{Subhradeep Kayal
}


\institute{S. Kayal \at
              Espoo, Finland \\
              \email{subhradeep.nitr@gmail.com}           
}

\date{Received: date / Accepted: date}

\maketitle

\begin{abstract}
Unsupervised image segmentation algorithms aim at identifying disjoint homogeneous regions in an image, and have been subject to considerable attention in the machine vision community. In this paper, a popular theoretical model with it's origins in statistical physics and social dynamics, known as the Deffuant-Weisbuch model, is applied to the image segmentation problem.
\\
The Deffuant-Weisbuch model has been found to be useful in modelling the evolution of a closed system of interacting agents characterised by their opinions or beliefs, leading to the formation of clusters of agents who share a similar opinion or belief at steady state. In the context of image segmentation, this paper considers a pixel as an agent and it's colour property as it's opinion, with opinion updates as per the Deffuant-Weisbuch model. Apart from applying the basic model to image segmentation, this paper incorporates adjacency and neighbourhood information in the model, which factors in the local similarity and smoothness properties of images. Convergence is reached when the number of unique pixel opinions, i.e., the number of colour centres, matches the pre-specified number of clusters.
\\
Experiments are performed on a set of images from the Berkeley Image Segmentation Dataset and the results are analysed both qualitatively and quantitatively, which indicate that this simple and intuitive method is promising for image segmentation. To the best of the knowledge of the author, this is the first work where a theoretical model from statistical physics and social dynamics has been successfully applied to image processing.

\end{abstract}

\section{Introduction}
\label{intro}
Image segmentation is the process of dividing an image into meaningful regions. For a segmentation technique to be useful for image analysis and interpretation, the separated regions should strongly relate to depicted objects or features of interest. Image segmentation is an important step in various image processing tasks as it transforms a low-level image to high-level image descriptions, in terms of features, objects, and scenes. A wide variety of methods exist in literature, with most of the segmentation algorithms belonging to one of the following broad categories: threshold based, edge-detection based, region based or based on clustering techniques. \cite{reviewpaper1} and \cite{reviewpaper2} provide excellent reviews of existing state-of-the-art image segmentation techniques and applications.

In this paper, a theoretical model from statistical physics, which aims at studying the properties of a population of interacting agents, is applied to the problem of image segmentation. The model described in this paper, and such other models, were originally motivated by the interaction of molecules in fluids, and have been used widely in studying social dynamics. This is the first cross-disciplinary practical use of such a model, according to the best of the knowledge of the author. Apart from using the original model, modifications are also suggested to utilise the spatial and neighbourhood information within the images to impose smoothness.

\section{Paper Structure}

The rest of the paper is organised as follows: the following Section \ref{opinion} provides some background of concepts from statistical physics and introduces the Deffuant-Weisbuch model, which is the central model of this paper. Modifications to the model are proposed in Section \ref{main}, to include spatial and neighbourhood information. Section \ref{exp} outlines the experiments performed and their results, and Section \ref{conclusion} concludes the paper and lists some possible future directions.

\section{Statistical Physics and Social Dynamics}
\label{opinion}

The development of the kinetic theory of gases, which sought to describe a system by focusing, not on a single particle, but on the system as a whole, gave rise to the field of modern statistical physics \cite{StatPhys}. Since then, statistical physics has been applied to fields as diverse as medicine, computer science and economics among others \cite{StatPhysExample}.

One important area of application of the modelling techniques from statistical physics is in the investigation of the dynamics in social phenomena \cite{sociophysics}, where the end goal is to understand the large-scale effects of collective interaction between agents. Each individual, in such a setting, is modelled as a simple entity having an opinion or a set of opinions, and interaction between agents leads to change in these opinions over time. This assumption, although rather simplistic, has been found to be quite robust in large-scale settings \cite{socialatom}. With this assumption, the aim is to study the steady opinion states of a population, and the processes that determine the interactions.

While there are many models which have been used to study opinion-change dynamics, the following subsection briefly summarises a model which has been used in this paper for the purpose of image segmentation. For a more in-depth and comprehensive review of models from statistical physics and their use in social dynamics, the papers \cite{SocialDynamicsReview} and \cite{DeffuantOverview} may be consulted.

\subsection{The Deffuant-Weisbuch Model of Consensus Formation}
\label{deffuant}
The basic model developed in \cite{deffuant} considers a population of \begin{math} N \end{math} agents \begin{math} i \end{math} with continuous opinions \begin{math} x_t^i \end{math} at time \begin{math} t \end{math}. Two randomly chosen agents meet at every time step and interaction happens if there opinions differ by less than a certain threshold \begin{math} \epsilon \end{math}, causing them to affect each other's opinion. Thus, if two agents \begin{math} i \end{math} and \begin{math} j \end{math} have opinions \begin{math} x_t^i \end{math} and \begin{math} x_t^j \end{math}, and the difference in opinion \begin{math} | x_t^i - x_t^j | < \epsilon \end{math}, then the opinions for the next time step are adjusted according to:
\begin{equation}
\begin{aligned}
x_{t+1}^i = x_t^i + \mu (x_t^j - x_t^i) \\
x_{t+1}^j = x_t^j + \mu (x_t^i - x_t^j)
\end{aligned}
\label{updateeq}
\end{equation}
where, \begin{math} \mu \end{math} is the convergence parameter, which varies between 0 and 0.5.

The reason for the imposition of such a threshold condition is that, the agents only interact with each other if their opinions are 'close enough'. In that case, their opinions will symmetrically get closer to each other, the final result being one or more clusters, depending on the value of the confidence threshold \begin{math} \epsilon \end{math} (as shown by \cite{clustersconfidence}, number of clusters \begin{math} c \end{math} is given by \begin{math} c \approx \lfloor \frac{1}{2 \epsilon} \rfloor \end{math}).

\begin{figure}[!htbp]
\centering
\includegraphics[width=1.0\linewidth]{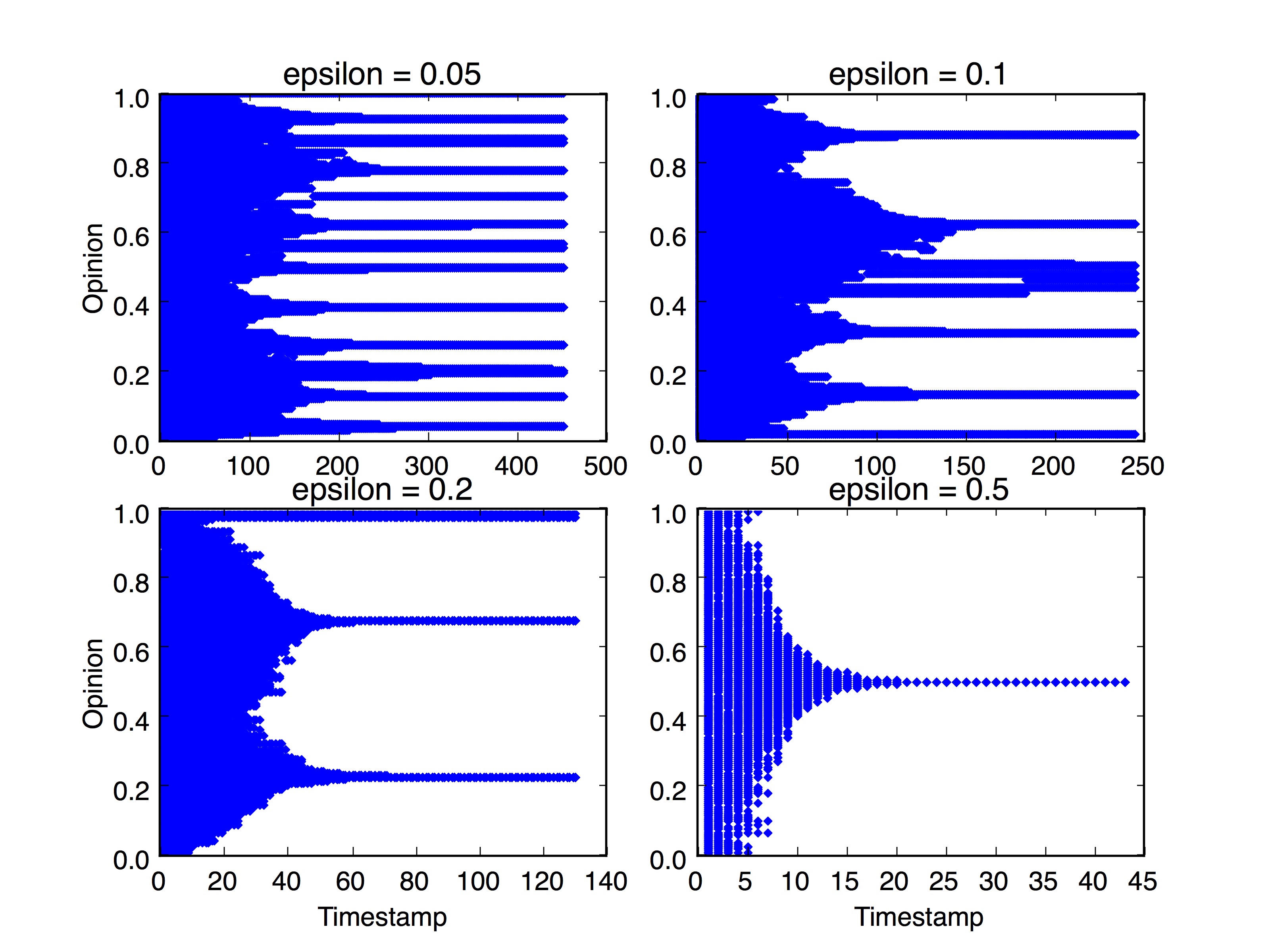}
\caption{Evolution of opinions for different values of the confidence threshold parameter.}
\label{deffuantfigure}
\end{figure}

To illustrate, Figure \ref{deffuantfigure} shows the results of computer simulations of the evolution of opinions, all of which were initialised with \begin{math} N = 1000 \end{math} agents with opinions distributed uniformly between 0 and 1.

The Deffuant-Weisbuch model is subject to two explicit parameters, the confidence threshold \begin{math} \epsilon \end{math} and the convergence parameter \begin{math} \mu \end{math}. While \begin{math} \mu \end{math} mainly influences the convergence time of the model by changing the size of the update, \begin{math} \epsilon \end{math} is the main factor determining opinion convergence and stability. Studies, such as \cite{fortunato}, show that there exists a critical value for the confidence threshold above which the agents' opinions can reach consensus, and below which consensus is difficult. Researchers have also found that the critical value for the confidence threshold is related to the initial opinion distribution \cite{hirscher} \cite{shang}.

\subsection{Why the Deffuant-Weisbuch Model?}
As mentioned earlier, there are many models in the social dynamics literature, out of which the Deffuant-Weisbuch model was chosen for the image segmentation experiments. The rationale behind this is the fact that the models which were suggested before the Deffuant-Weisbuch model, namely, the Ising Model \cite{ising}, the Voter Model \cite{voter}, the Majority Rule Model \cite{majority}, the Sznajd Model \cite{sznajd}, are too primitive and/or work on only discrete opinions, whereas a contemporary of this model, the Hegselmann-Krause Model \cite{hkmodel}, has a rather long running time to be meaningful in this context.

\section{Modifications to the Deffuant-Weisbuch Model and Application to Image Segmentation}
\label{main}

\subsection{Application to Image Segmentation}
An image in it's basic form is a grid of pixels, each of which has a colour property which might be a vector or a scalar. Treating each pixel as an agent, such that it's colour property is it's opinion, a natural setting to apply the Deffuant-Weisbuch model is arrived at. If the image is grayscale, then the opinion of the pixel agent is a single number and the model can be readily applied, whereas if the pixel value is a vector, i.e., RGB, HSV, YCbCr etc, then the update rule is applied to interacting pixel agents if:
\begin{equation}
\max ( | \vec{x_t^i} - \vec{x_t^j} | ) < \epsilon
\label{condition}
\end{equation}

Upon convergence, the image will be expected to have a small number of colour centres to which the pixel values will have converged to, similar to another popular algorithm, the K-means \cite{kmeans}. The number of the final colour centres depends on the confidence threshold parameter \begin{math} \epsilon \end{math}.

Having discussed about applying the Deffuant-Weisbuch model in the scenario of image segmentation, the next two subsections focus on the following:
\begin{enumerate}
\item
In Section \ref{neighbourhood}, the model is modified to include the neighbourhood information around each pixel, which impose smoothness constraints.
\item
In a typical clustering problem, the parameter to be chosen is the number of clusters \begin{math} c \end{math}. However, in the Deffuant-Weisbuch model, the number of clusters is implicitly determined by the choice of the confidence threshold \begin{math} \epsilon \end{math}, which is difficult to select since it is only approximately related to the number of clusters (see Section \ref{deffuant}). In Section \ref{clustering}, a simple solution to this problem is suggested.
\end{enumerate}

\subsection{Adding Neighbour Information to the Deffuant-Weisbuch Model}
\label{neighbourhood}

The Deffuant-Weisbuch model works with interactions among individual agents, and does not take into consideration any information about the spatial neighbourhood of the agent. Since, in an image, spatially adjacent pixels presumably have a stronger connection than pixels which are far away from each other, it is important to take into account the local commonality of location while updating the opinion of the pixel agents. In this paper, two different mechanisms of opinion-update with locational information have been experimented with.

\subsubsection{Using the Distance Between Interacting Agents}
\label{distancebased}

During the updating of the opinions in the Deffuant-Weisbuch model, as described in Equation \ref{updateeq}, two pixel agents \begin{math} i \end{math} and \begin{math} j \end{math} are chosen at random, and their opinions are updated if they satisfy Equation \ref{condition}. Let the coordinates of the pixels in the image be \begin{math} p_i \end{math} and \begin{math} p_j \end{math}, and the coordinates of the top-left and bottom-right corners of the image be \begin{math} p_0 \end{math} and \begin{math} p_{size} \end{math} respectively. Then the distance between them is calculated as the normalised minkowski distance:
\begin{equation}
d = \frac{\mathrm{minkowski} (p_i, p_j)}{\mathrm{minkowski}(p_0, p_{size})}
\label{distanceeq}
\end{equation}
where,
\begin{displaymath}
\mathrm{minkowski} (a, b) = {\left(\sum_{m=1}^{n} {| a_m - b_m |}^k\right)}^{1/k}
\end{displaymath}
for \begin{math} k > 1 \end{math}, and with \begin{math} m \end{math} as the dimensionality.

Using Equations \ref{updateeq} and \ref{distanceeq}, the new update rule becomes:
\begin{equation}
\begin{aligned}
x_{t+1}^i = x_t^i + \mu (x_t^j - x_t^i) (1 - d) \\
x_{t+1}^j = x_t^j + \mu (x_t^i - x_t^j) (1 - d)
\end{aligned}
\label{updateeq1}
\end{equation}

\subsubsection{Using the Neighbourhood of Interacting Agents}
\label{neighbourhoodbased}

Instead of using the opinion of a pixel agent, during the opinion update, the average opinion of 'like-minded' agents in the neighbourhood of that pixel agent in used.

The average neighbourhood opinion is given by:
\begin{equation}
\eta (x_t^i) = \sum_{y_t \in \mathcal{N}_{x_t^i}} \mathcal{I} y_t
\end{equation}
where, \begin{math} \mathcal{N}_{x_t^i} \end{math} denotes the 4 or 8-neighbourhood of \begin{math} x_t^i \end{math} including the point itself, and \begin{math} \mathcal{I} \end{math} is an binary indicator variable which is 1 when the condition given in Equation \ref{condition} is met for  \begin{math} y_t \end{math} and  \begin{math} x_t^i \end{math}, and 0 otherwise. Then using the above equation in the update, the following is arrived at:
\begin{equation}
\begin{aligned}
x_{t+1}^i = \eta (x_t^i) + \mu (\eta (x_t^j) - \eta (x_t^i)) \\
x_{t+1}^j = \eta (x_t^j) + \mu (\eta (x_t^i) - \eta (x_t^j))
\end{aligned}
\label{updateeq2}
\end{equation}

\subsection{Iterative Adjustment of the Confidence Threshold $\epsilon$}
\label{clustering}

In the Deffuant-Weisbuch model, the number of opinion clusters after convergence is determined by the confidence threshold parameter \begin{math} \epsilon \end{math}. In the work of \cite{clustersconfidence}, it was proved that \begin{math} c \approx \lfloor \frac{1}{2 \epsilon} \rfloor \end{math}, which is not an exact relationship between the number of clusters and the confidence threshold, thereby making an initial selection of \begin{math} \epsilon \end{math} difficult.

In order to tackle this problem, the process of clustering is started with a small initial value of \begin{math} \epsilon \end{math}, which is increased slowly with every run of the model, until the number of clusters at convergence matches a pre-specified one.

\subsection{Final Algorithms}
With sufficient background knowledge, the algorithm pseudo-codes are stated next.
\begin{itemize}
\item
Algorithm \ref{algo1} outlines the steps required to implement the Deffuant-Weisbuch model.
\item
Algorithm \ref{algo2} builds on Algorithm \ref{algo1} to adjust the value of confidence parameter to segment the image into a given number of clusters.
\end{itemize}

\begin{algorithm}[!htbp]
 \KwData{Image, confidence threshold \begin{math} \epsilon \end{math}}
 \KwResult{Image with \begin{math} c \end{math} colour centres}
Let the image at time \begin{math} t \end{math} be denoted by \begin{math} I_t \end{math}, and the pixels at position \begin{math} p_i \end{math} and \begin{math} p_j \end{math} have values as \begin{math} x_t^i \end{math} and \begin{math} x_t^j \end{math}. Let \begin{math} \mu = 0.5 \end{math} and the convergence criterion \begin{math} e = 10^{-6} \end{math}\;
    \Do{\begin{math} \mathrm{diff_t} > e \end{math}}{
      \begin{math} x_{t+1}^i = \mathrm{update} (x_t^i, x_t^j, \mu)\end{math} and \begin{math} x_{t+1}^j = \mathrm{update} (x_t^i, x_t^j, \mu) \end{math}, where update uses one of Equations \ref{updateeq}, \ref{updateeq1} or \ref{updateeq2}\;
      \begin{math} \mathrm{diff_t} = \max \left( \mathrm{abs} (I_t - I_{t-1}) \right)\end{math}\;
    }
\caption{The Deffuant-Weisbuch Model}
\label{algo1}
\end{algorithm}

\begin{algorithm}[!htbp]
 \KwData{Image \begin{math} I_0 \end{math}, the final number of clusters \begin{math} c \end{math}, initial confidence threshold \begin{math} \epsilon_0 \end{math} and confidence threshold increment \begin{math} \Delta \epsilon \end{math}}
 \KwResult{Image with \begin{math} c \end{math} colour centres}
Let the image at iteration \begin{math} i \end{math} be denoted by \begin{math} I_i \end{math}, while the confidence parameter is denoted by \begin{math} \epsilon_i \end{math}. Let \begin{math} \mathrm{count\_clusters} (I) \end{math} be a function which takes in an image and counts the number of cluster centres in the image. The starting conditions are \begin{math} I_i = I_0 \end{math} and \begin{math} \epsilon_i = \epsilon_0 \end{math}\;
    \Do{\begin{math} c_i < c \end{math}}{
      \begin{math} I_{i} = \mathrm{deffuant} (I_{i-1}, \epsilon_{i-1})\end{math}\;
      \begin{math} \epsilon_{i} = \epsilon_{i-1} + \Delta \epsilon \end{math}\;
      \begin{math} c_{i} = \mathrm{count\_clusters} (I_{i}) \end{math}\;
    }
\caption{Iterative Updating of Confidence Threshold}
\label{algo2}
\end{algorithm}

\section{Experiments}
\label{exp}

\subsection{Dataset}
In order to provide a quantitative evaluation of this method, experiments are conducted on a small subset of 25 images extracted from the Berkeley Image Segmentation Dataset \cite{Berkeley}, which contain meaningful object and background entities. Apart from the natural images which are used in the segmentation process as input, the dataset also contains manual border annotation of the images, which are considered the ground truth.

\subsection{Pre and Post-Processing}
Pre and post-processing is done in all the experiments. In order to reduce noise, the input images were subjected to bilateral filtering \cite{bilateral}, which smoothes images but preserves edges. Also, the images after clustering are subjected to morphological smoothing operations to remove small isolated components.

\subsection{Features}
Since the emphasis of this paper is on the usefulness of the Deffuant-Weisbuch model, complex high-level pixel features have not been extracted to be used in the segmentation process. The raw RGB value of each pixel is used as the opinion vector for the clustering process.

\subsection{Benchmarks}

\subsubsection{K-Means Clustering}
The results of the Deffuant-Weisbuch scheme are compared with those of the K-Means Clustering algorithm \cite{kmeans}. The K-means algorithm is chosen as a benchmark because it is a popular well-established machine learning algorithm \cite{topten}, and as it also clusters based on the distance between the points.

\subsubsection{Simple Linear Iterative Clustering}
The Simple Linear Iterative Clustering algorithm (or SLIC in short) is a special case of the K-Means algorithm, specific to superpixel segmentation \cite{slic1}. SLIC clusters pixels based on their coordinates in a five-dimensional space, three of which are the colour coordinates in the CIELAB space, and two are the pixel position coordinates in the image. As shown in \cite{slic2}, SLIC can be considered quite state-of-the-art.

\begin{figure}[!htbp]
\subfloat{\includegraphics[width=0.33\linewidth]{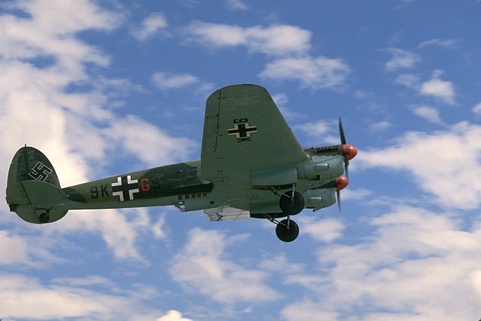}}
\setcounter{subfigure}{0}
\subfloat[Original]{\includegraphics[width=0.33\linewidth]{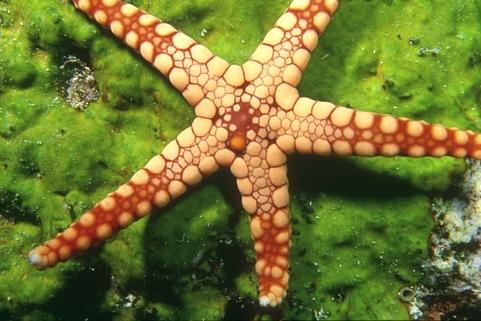}}
\subfloat{\includegraphics[width=0.33\linewidth]{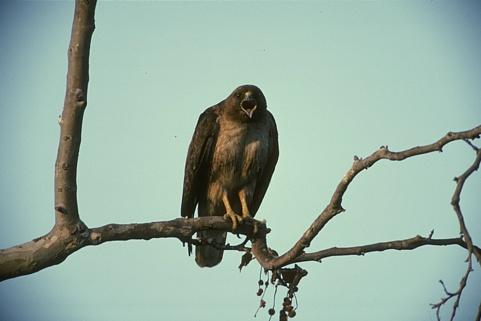}}\\
\subfloat{\includegraphics[width=0.33\linewidth]{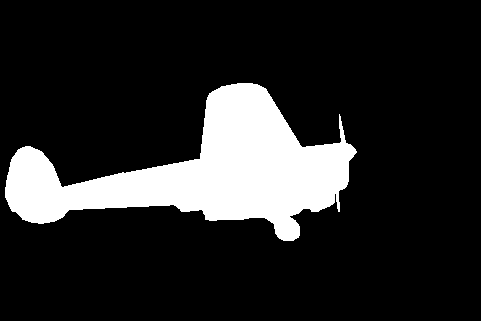}}
\setcounter{subfigure}{1}
\subfloat[Ground Truth]{\includegraphics[width=0.33\linewidth]{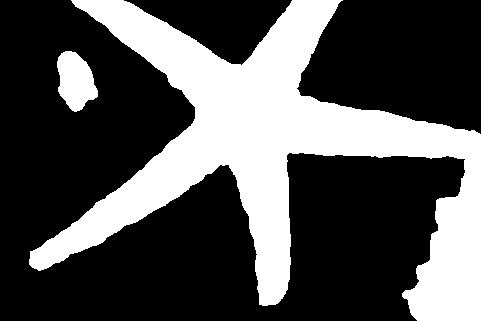}}
\subfloat{\includegraphics[width=0.33\linewidth]{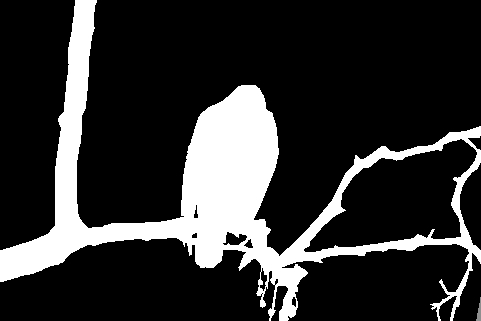}}\\
\subfloat{\includegraphics[width=0.33\linewidth]{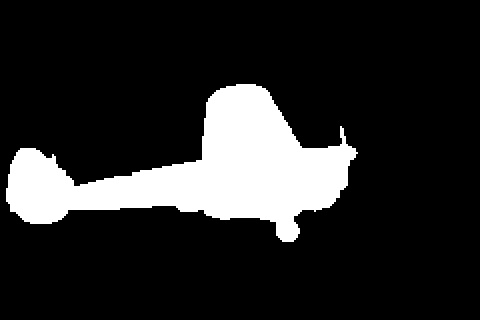}}
\setcounter{subfigure}{2}
\subfloat[Deffuant-Neighbour]{\includegraphics[width=0.33\linewidth]{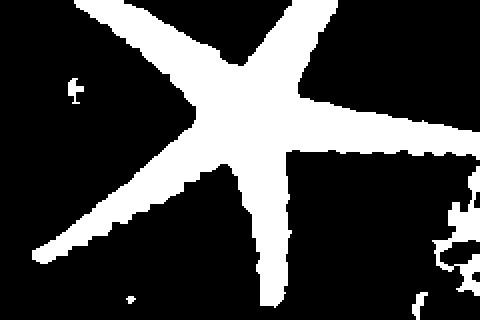}}
\subfloat{\includegraphics[width=0.33\linewidth]{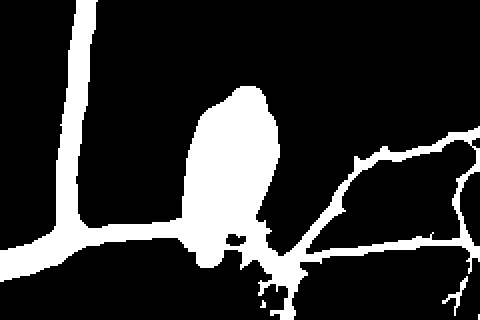}}\\
\subfloat{\includegraphics[width=0.33\linewidth]{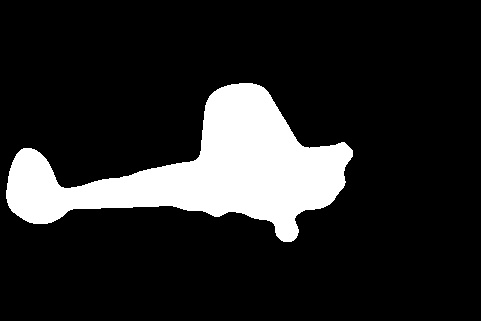}}
\setcounter{subfigure}{3}
\subfloat[SLIC]{\includegraphics[width=0.33\linewidth]{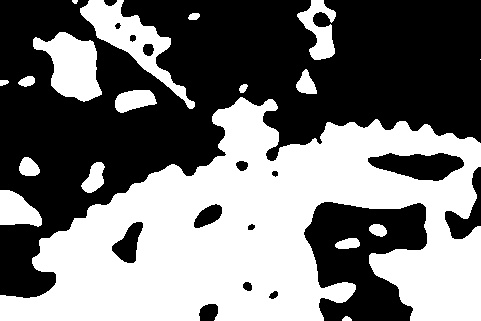}}
\subfloat{\includegraphics[width=0.33\linewidth]{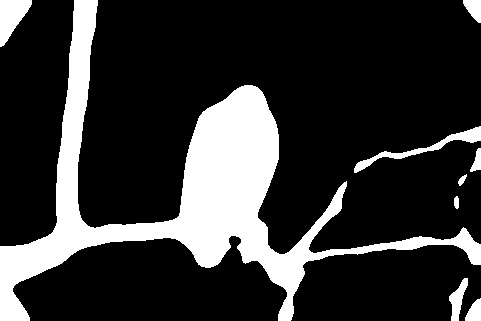}}\\
\subfloat{\includegraphics[width=0.33\linewidth]{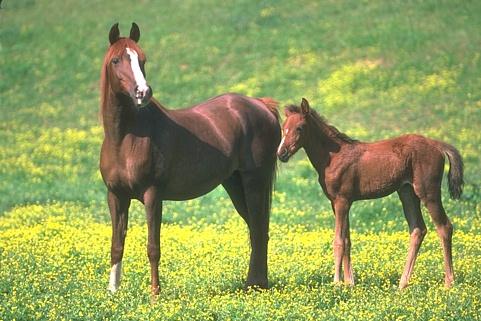}}
\setcounter{subfigure}{0}
\subfloat[Original]{\includegraphics[width=0.33\linewidth]{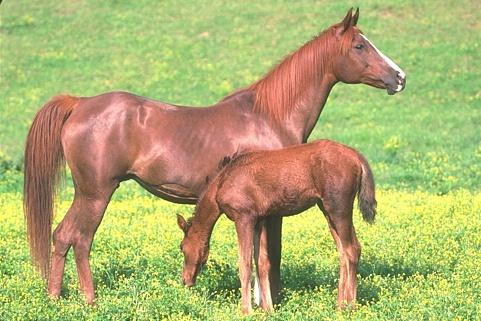}}
\subfloat{\includegraphics[width=0.33\linewidth]{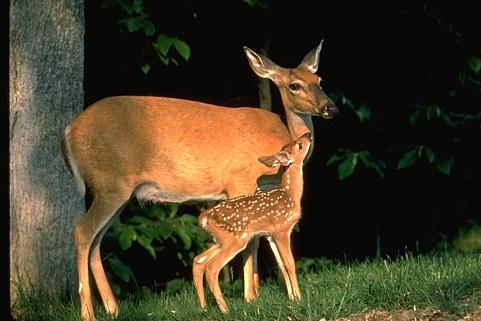}}\\
\subfloat{\includegraphics[width=0.33\linewidth]{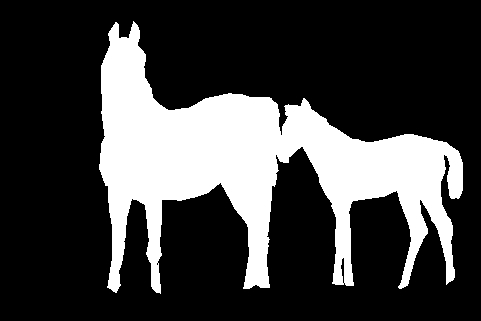}}
\setcounter{subfigure}{1}
\subfloat[Ground Truth]{\includegraphics[width=0.33\linewidth]{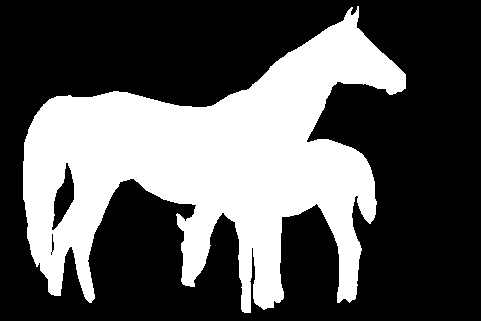}}
\subfloat{\includegraphics[width=0.33\linewidth]{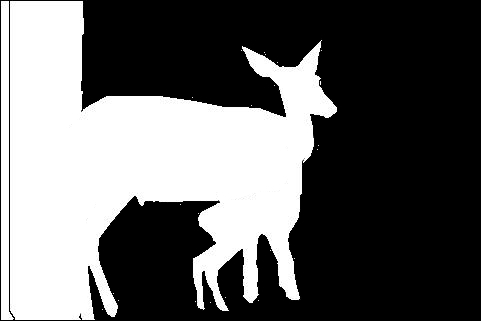}}\\
\subfloat{\includegraphics[width=0.33\linewidth]{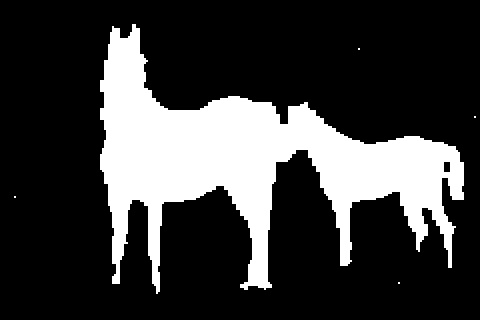}}
\setcounter{subfigure}{2}
\subfloat[Deffuant-Neighbour]{\includegraphics[width=0.33\linewidth]{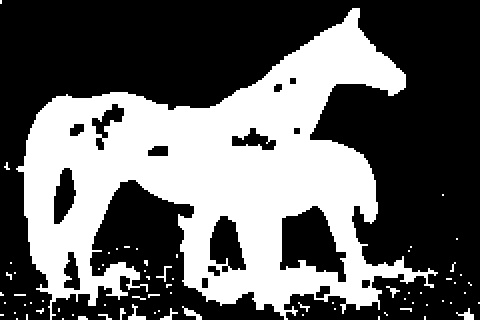}}
\subfloat{\includegraphics[width=0.33\linewidth]{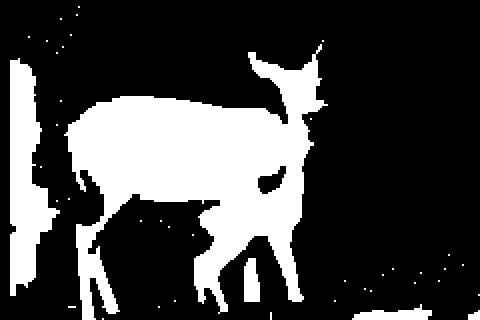}}\\
\subfloat{\includegraphics[width=0.33\linewidth]{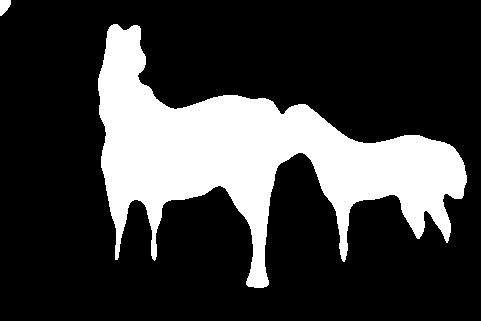}}
\setcounter{subfigure}{3}
\subfloat[SLIC]{\includegraphics[width=0.33\linewidth]{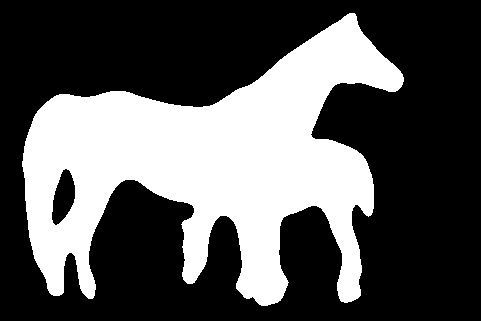}}
\subfloat{\includegraphics[width=0.33\linewidth]{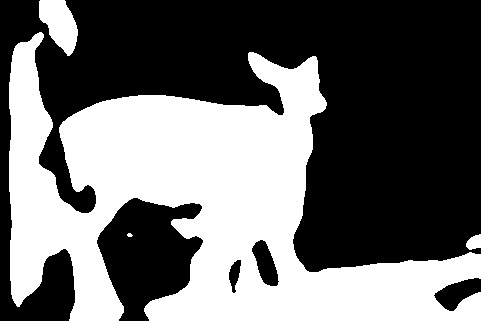}}
\caption{Example segmentation results with c=2 : (a) Original Images, (b) Ground Truth Segmentation, (c) Segmentation with Deffuant-Weisbuch Model with Neighbourhood information, (d) Segmentation with the SLIC algorithm }
\label{results}
\end{figure}

\subsection{Evaluation}
The evaluation measures used in this paper are suggested in \cite{eval}. First, the number of True Positive \begin{math} TP \end{math} (classified as object by both the algorithm and the ground truth), True Negative \begin{math} TN \end{math} (classified as non-object by both the algorithm and the ground truth), False Positive \begin{math} FP \end{math} (non-object classified as object by both the algorithm and the ground truth) and False Negative \begin{math} FN \end{math} (object classified as non-object by both the algorithm and the ground truth) pixels were counted. Then, the recall, fallout and accuracy are calculated as:

\begin{gather}
\mathrm{accuracy} = \frac{TP + TN}{TP + FP + TN + FN} \nonumber \\
\mathrm{recall} = \frac{TP}{TP + FN} \\
\mathrm{fallout} = \frac{FP}{FP + TN} \nonumber
\end{gather}

\subsection{Results}

Experiments are run with \begin{math} c = 2 \end{math}, initial confidence threshold \begin{math} \epsilon_0 = 0.1 \end{math} and threshold increment \begin{math} \Delta \epsilon = 0.01 \end{math}. The results are stated in the following table, where the algorithm name 'Deffuant' denotes the standard Deffuant-Weisbuch model (with the opinion updates given by Equation \ref{updateeq}), 'Deffuant-Distance' refers to the modified model with distance information (using update Equation \ref{updateeq1}), and 'Deffuant-Neighbour' is the modified model with neighbourhood information (using update Equation \ref{updateeq2}).

\begin{table}[!htbp]
\centering
\caption{Segmentation Results}
  \begin{tabular}{ | c || c | c | c | }
    \hline
    \textbf{Name} & \textbf{Recall} & \textbf{Fallout} & \textbf{Accuracy}\\ \hline
    \textbf{K-Means} & 73.49\% & 11.74\% & 85.98\%\\ \hline
    \textbf{SLIC} & \textbf{79.64\%} & 9.11\% & 88.61\%\\ \hline
    \textbf{Deffuant} & 72.54\% & 3.03\% & 92.27\%\\ \hline
    \textbf{Deffuant-Distance} & 76.78\% & 3.00\% & 92.55\%\\ \hline
    \textbf{Deffuant-Neighbour} & 78.28\% & \textbf{2.89\%} & \textbf{92.97\%}\\ \hline
  \end{tabular}
  \label{resultstable}
\end{table}

Bearing in mind that high values of accuracy and recall are better, while a low value of fallout is desired, it can be quickly observed that even the standard Deffuant-Weisbuch model achieves respectable scores relative to both the K-Means and the SLIC algorithm. The modifications to the model bring in further improvements, especially to the recall values, which means that the model gets better at correctly identifying the object pixels.

Apart from the results in Table \ref{resultstable}, some qualitative results of segmentation are shown in Figure \ref{results}.

\section{Conclusions and Future Work}
\label{conclusion}

In this paper, a popular theoretical model from statistical physics, the Deffuant-Weisbuch model, is used for unsupervised image segmentation, with two different modifications to include spatial information. Quantitative evaluation of the algorithm is done by using it to segment images from a small dataset into 2 clusters, and is compared to the results of the well-known K-Means algorithm and the state-of-the-art SLIC algorithm. Results suggest good performance, especially in the algorithm's ability to correctly identify object pixels in this 2-cluster setting.

Possible future work includes testing the performance of the Deffuant-Weisbuch model and it's modifications on images from different domains, such as hyper-spectral and medical images, and more sophisticated use of neighbourhood information.


\bibliographystyle{spmpsci}      

\bibliography{refs}

\end{document}